\documentclass[runningheads]{llncs}
\usepackage[T1]{fontenc}
\usepackage{graphicx}
\usepackage{subfig}
\usepackage{multirow}
\usepackage{amsmath}
\usepackage{amssymb}
\usepackage{booktabs}
\usepackage{epsfig}
\usepackage{bbding}
\usepackage{float}
\usepackage{multicol}
\usepackage[colorlinks,
linkcolor=blue,
anchorcolor=blue,
citecolor=blue,
urlcolor=blue]{hyperref}
\usepackage[table]{xcolor}
\usepackage{color}
\usepackage[table]{xcolor}
\definecolor{mygray}{RGB}{230, 230, 230}

\begin{document}
\title{Surgformer: Surgical Transformer with Hierarchical Temporal Attention for Surgical Phase Recognition}
\titlerunning{Surgformer}
%
\author{Shu Yang\inst{1}\orcidID{0000-0002-1761-9286} 
\and
Luyang Luo\inst{1}\orcidID{0000-0002-7485-4151} 
\and
Qiong Wang\inst{2}\orcidID{0000-0001-9515-7745} 
\and
Hao Chen\thanks{Corresponding Author}\inst{1,3,4}\orcidID{0000-0002-8400-3780} 
}
\authorrunning{S. Yang et al.}
%
\institute{Department of Computer Science and Engineering, The Hong Kong
University of Science and Technology, Hong Kong, China \and
Guangdong Provincial Key Laboratory of Computer Vision and Virtual Reality Technology, Shenzhen Institute of Advanced Technology, Chinese Academy of Sciences  \and
Department of  Chemical and Biological Engineering, The Hong Kong
University of Science and Technology, Hong Kong, China \and
HKUST Shenzhen-Hong Kong Collaborative Innovation Research Institute, Futian, Shenzhen, China\\
\email{syangcw@connect.ust.hk}, \email{jhc@cse.ust.hk}}
\maketitle              
\begin{abstract}
Existing state-of-the-art methods for surgical phase recognition either rely on the extraction of spatial-temporal features at a short-range temporal resolution or adopt the sequential extraction of the spatial and temporal features across the entire temporal resolution. However, these methods have limitations in modeling spatial-temporal dependency and addressing spatial-temporal redundancy: 1) These methods fail to effectively model spatial-temporal dependency, due to the lack of long-range information or joint spatial-temporal modeling. 2) These methods utilize dense spatial features across the entire temporal resolution, resulting in significant spatial-temporal redundancy. In this paper, we propose the Surgical Transformer (Surgformer) to address the issues of spatial-temporal modeling and redundancy in an end-to-end manner, which employs divided spatial-temporal attention and takes a limited set of sparse frames as input. Moreover, we propose a novel Hierarchical Temporal Attention (HTA) to capture both global and local information within varied temporal resolutions from a target frame-centric perspective. Distinct from conventional temporal attention that primarily emphasizes dense long-range similarity, HTA not only captures long-term information but also considers local latent consistency among informative frames. HTA then employs pyramid feature aggregation to effectively utilize temporal information across diverse temporal resolutions, thereby enhancing the overall temporal representation. Extensive experiments on two challenging benchmark datasets verify that our proposed Surgformer performs favorably against the state-of-the-art methods. The code is released at \url{https://github.com/isyangshu/Surgformer}.

\end{abstract}
\begin{figure}[t]
\centering
\includegraphics[width=1.0\linewidth]{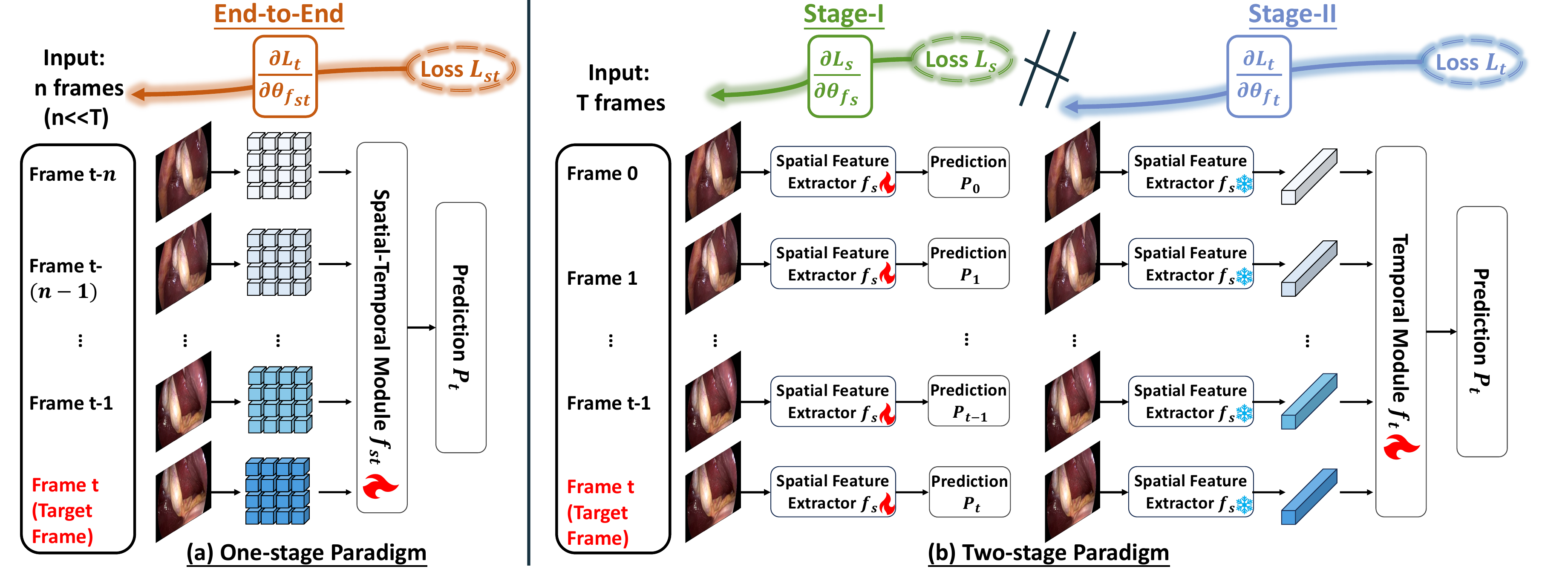}
\caption{Illustration of architectures of different paradigms. {\bf left:}  the one-stage paradigm with a unified spatial-temporal feature extractor. {\bf right:} the two-stage paradigm with separate spatial and temporal feature extractors.}
\label{fig:motivation}
\end{figure}

\section{Introduction}
Surgical phase recognition aims to automatically allocate the workflow categories of surgery for each frame in a surgical video, which enhances the quality of computer-assisted interventions and facilitates numerous potential medical applications, such as monitoring of surgical procedures~\cite{OR1,OR2}, surgical competence assessment~\cite{skill_assessment2,skill_assessment1}, and surgical workflow optimization~\cite{optimization}. 

An intuitive solution, termed as the one-stage paradigm, is to directly extract spatial-temporal features from a confined temporal window with target frame in an end-to-end manner. As depicted in Fig.~\ref{fig:motivation} (a), the one-stage paradigm employs a limited subset of frames as input to avoid the huge computational overhead. Existing one-stage methods utilize conventional temporal modules, such as LSTM~\cite{SV-RCNet,MTRCNet-CL,TMRNet}, to facilitate spatial-temporal modeling. Nevertheless, the one-stage methods utilize conventional temporal modules on short-range temporal resolution, thereby struggling to effectively model long-term spatial-temporal dependency. An alternative solution is the two-stage paradigm, shown in the Fig.~\ref{fig:motivation} (b), which sequentially extracts spatial and temporal features over the entire temporal resolution. Recent advances have been primarily driven by two-stage methods~\cite{TeCNO,Trans-SVNet,LoViT,SKiT}. Specifically, the two-stage paradigm initially converts the target frame and all preceding frames into spatial features, followed by transforming these spatial features into spatial-temporal features via additional temporal modules. Note that the gradient flow is interrupted between two stages due to significant training overhead of joint backpropagation. Despite the impressive performance, existing approaches suffer from limitations in enabling comprehensive spatial-temporal dependency modeling, since the reliance on two-stage pipeline impedes the concurrent modeling of temporal and spatial feature representations. Moreover, both two paradigms utilize all the spatial features throughout the temporal resolution, leading to noticeable redundancy in spatial-temporal information.

We argue that concurrent spatial-temporal modeling is essential for frame-level tasks~\cite{AMCNet,FVOS,UVOSHT,MATNet}, and hierarchical integration of long-term and short-term representations~\cite{slowfast,PhiTrans} is beneficial for the understanding of spatial-temporal information. And a limited set of sparse frames can yield adequate spatial-temporal representations due to the subtle target motions observed across adjacent frames. Thus, we reconsider the one-stage paradigm with a spatial-temporal transformer, which can naturally tackle the aforementioned challenges associated with spatial-temporal modeling and redundancy. Nevertheless, existing transformers~\cite{ViViT,TimeSFormer} tailored for video recognition prioritize global consistency and neglect the potential contribution of target frame, which is important for surgical phase recognition.

\begin{figure}[t]
\centering
\includegraphics[width=0.80\linewidth]{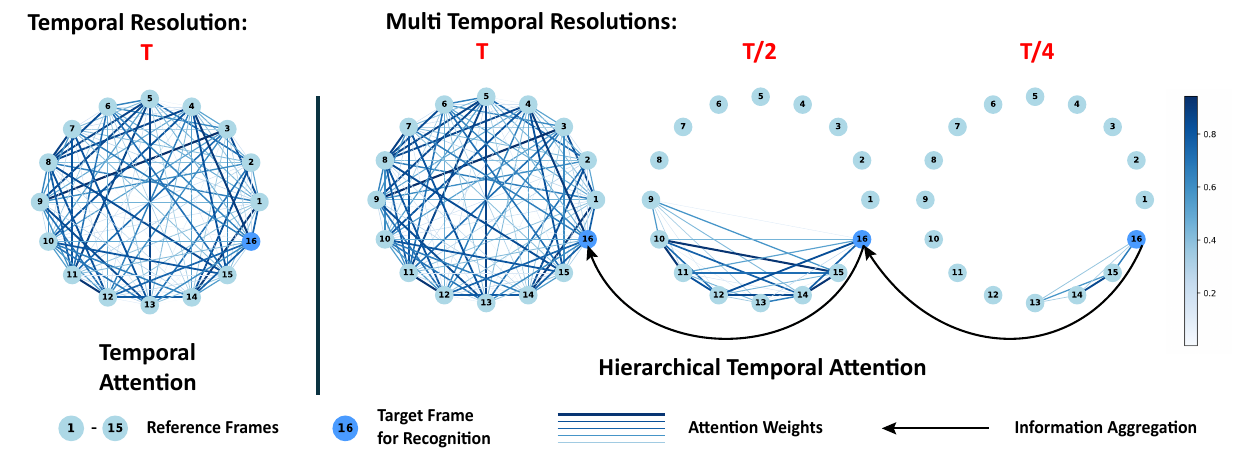}
\caption{Illustration of temporal attention mechanisms. In contrast to temporal attention only establishing dense global dependencies among frames, Hierarchical Temporal Attention constructs multiple temporal segments with different temporal resolutions to capture the long-term and short-term information.}
\label{fig:attention}
\end{figure}

Motivated by the above observations, we propose the \textbf{Surg}ical Trans\textbf{former} (\textbf{Surgformer}) with the following contributions: (1) Surgformer employs divided spatial-temporal attention~\cite{TimeSFormer} and takes a limited set of sparse frames as input to address the issues of spatial-temporal modeling and redundancy in an end-to-end manner. (2) We propose the Hierarchical Temporal Attention (HTA) (shown in Fig.~\ref{fig:attention}) to capture both global and local information from a target frame-centric perspective. HTA partitions the temporal sequence into multiple temporal segments of distinct temporal resolutions, and flexibly establishes token dependencies within varied-range temporal segments. Subsequently, HTA employs pyramid feature aggregation to leverage temporal tokens across varied temporal resolutions. (3) To investigate the effectiveness of our proposed model, we conduct comprehensive experiments including overall comparison and ablation studies on two challenging datasets, which demonstrates that our proposed method performs favorably against the state-of-the-art methods.

\section{Method}
\begin{figure}[t]
\centering
\includegraphics[width=0.9\linewidth]{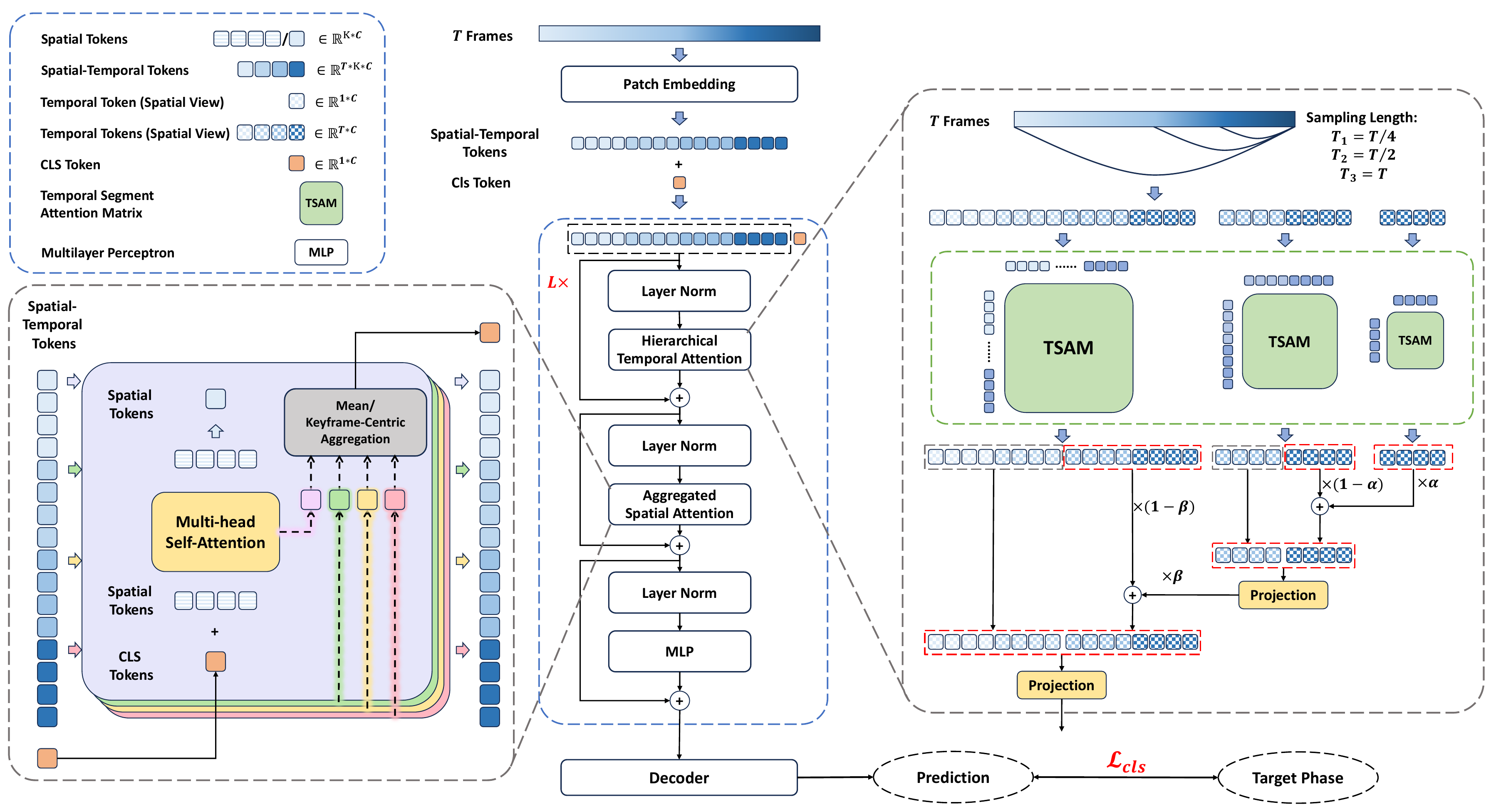}
\caption{Overview of Surgformer. Given spatial-temporal tokens, we sequentially utilize Hierarchical Temporal Attention and Aggregated Spatial Attention to  facilitate the learning of spatial-temporal feature representations.}
\label{fig:method}
\end{figure}
\subsection{Overview of Surgformer}

Given untrimmed surgical footage, we sample $T$ frames forward starting from the target frame to generate the frame volume $V\in\mathbb{R}^{T\times C\times H\times W}$, with the objective of predicting the corresponding surgical phase. For the frame volume $V$, $T$ denotes the temporal resolution, $C$ is the number of channel, and $(H,W)$ is the spatial resolution of each frame. Meanwhile, due to the subtle target motions across adjacent frames, we employ the sparsification strategy to sample one frame from every $R$ frames, and obtain a total of $T$ frames.
The frame volume $V$ achieves a sufficient length to encompass crucial phase transitions, while concurrently mitigating spatial-temporal redundancy between adjacent frames. Subsequently, the frame volume $V$ is partitioned into non-overlapping patches $\in\mathbb{R}^{P\times P}$ arranged in spatial-temporal order, followed by mapping these patches to spatial-temporal tokens $X\in\mathbb{R}^{(T\times K)\times C}$. $T$ refers to the temporal resolution, corresponding to the temporal positions. In addition, $(P,P)$ is the spatial resolution of each image patch, and $K=HW/P^2$ is the resulting number of spatial patches. Additionally, an extra class token $[CLS]\in\mathbb{R}^{1\times C}$ is appended to spatial-temporal tokens, responsible for aggregating global video information. After integrating dynamic learnable 3D position information into all the tokens, we can convert the frame volume $V$ into spatial-temporal tokens $X^{cls}\in\mathbb{R}^{(T\times K+1)\times C}$, which are then fed into sequentially stacked transformer blocks. In each transformer block, $X^{cls}$ is initially fed to the Hierarchical Temporal Attention (HTA) to aggregate the temporal information at each spatial position, followed by utilizing subsequent Aggregated Spatial Attention (ASA) to propagate learned temporal information to other spatial locations. The entire transformer block can be formulated as
\begin{equation}
    X_t = HTA(LayerNorm(X^{cls}))+ X^{cls},
\end{equation}
\begin{equation}
    X_{st} = ASA(LayerNorm(X_t)) + X_t,
\end{equation}
\begin{equation}
    X' = MLP(LayerNorm(X_{st})) + X_{st}.
\end{equation}
Lastly, the class token $[CLS]$ is introduced as input to the decoder for prediction.

\subsection{Hierarchical Temporal Attention}
Conventional temporal attention mechanisms primarily emphasize establishing extensive temporal dependencies across the entire temporal resolution, which suffers from difficulty in effectively utilizing the local information that is essential for surgical phase recognition. We reconsider the ignored critical visual attributes nearby target frames, and propose the Hierarchical Temporal Attention (depicted in the right of Fig.~\ref{fig:method}) to capture global information and local consistency among informative frames from a target frame-centric perspective. This perspective enables us to identify and leverage informative frames that exhibit valuable cues within the varied-range temporal sequence.

By partitioning the spatial-temporal tokens $X\in\mathbb{R}^{T\times K\times C}$ along the spatial dimension, we obtain temporal tokens $X_{temporal}\in\mathbb{R}^{T\times C}$ at each spatial position, which are subsequently aggregated to capture the relationships among tokens along the temporal dimension. To facilitate the learning of discriminative spatial-temporal representation, we utilize HTA to capture the dynamic relations among frames at varying temporal resolutions. From a target frame-centric perspective, we generate a set of $m$ structured temporal segments with varied temporal resolutions $\{T_s\}_{s=1}^m$ to encompass varied-range informative frames, allowing for the examination of potential local consistency. We compute Temporal Segment Attention Matrix to explore the token relationships along temporal dimension, and the score between two tokens can be formulated as:
\begin{equation}
{M^s(P_{(t_i,k)}, P_{(t_j,k)})} = \begin{cases} \frac{\exp({Q(P_{(t_i,k)})^TK(P_{(t_j,k)})})}{\sum_{t_j' \in T_s}\exp({Q(P_{(t_i,k)})^TK(P_{(t_j',k)})})},&{t_i, t_j \in T_s} \\ 
{0,}&{otherwise} \\ 
\end{cases}
\end{equation}
where $s$ represents the $s$-th temporal segment with temporal resolution $T_s$. $P_{(t_i,k)}$ and $P_{(t_j,k)}$ refer to the tokens located in the same spatial position but with distinct temporal positions. $Q(\cdot)$ and $K(\cdot)$ are two distinct linear projections. Subsequently, we can derive $m$ matrices to flexibly establish varied-range dependencies from a target frame-centric perspective, denoted as $\{M^s \in\mathbb{R}^{T_s\times T_s}\}_{s=1}^m$. To prevent the learning of information from irrelevant frames, we assign a score of 0 to tokens that do not belong to the same temporal segment. By utilizing the $m$ matrices to aggregate temporal information, each token attends to other tokens within both long-range and short-range temporal resolutions, which facilitates the learning of discriminative features from both global and local perspectives.

Then, we refine temporal context by leveraging temporal tokens in varied temporal segments with distinct temporal resolutions. Given multiple hierarchical enhanced temporal tokens, we fuse potential contribution of each token in varied temporal segments by pyramid feature aggregation. Specifically, we conduct aggregation operation for enhanced temporal tokens sequentially according to the size of the temporal resolution. For shared temporal positions, we formulate the aggregation operation as an element-wise weighted addition with hyper-parameters $\alpha$ and $\beta$. For incoherent temporal positions, we conduct concatenation for aggregation.

\subsection{Aggregated Spatial Attention}
Given spatial-temporal tokens $X\in\mathbb{R}^{T\times K\times C}$, we obtain spatial tokens $X_{spatial}\in\mathbb{R}^{K\times C}$ at each temporal position. As depicted in the left of Fig.~\ref{fig:method}, we replicate $[CLS]$ for each temporal position and generate the corresponding spatial tokens $X_{spatial}^{cls}\in\mathbb{R}^{(K+1)\times C}$, which are fed into Multi-head Self-Attention for enhancement. Subsequently, we utilize two distinct aggregation strategies to aggregate all the spatially enhanced $[CLS]$ tokens, namely Mean Aggregation (MA) and Target Frame-centric Aggregation (TFA). Specifically, MA is utilized to compute the average of all the $[CLS]$ tokens, enabling the extraction of informative cues that span the entire temporal sequence. Conversely, TFA involves the calculation of the $[CLS]$ similarity between the target frame and other frames, thereby learning the relative significance of each frame with respect to the target frame. We conduct the experiments about MA and TFA in $\S$~\ref{as}.

\begin{table*}[t]
	\centering
	\caption{Ablation analysis of our proposed Surgformer on two datasets. \textcolor{red}{Red} indicates the performance improvement compared to the Baseline.}
    \resizebox{9.0cm}{!}{
	\begin{tabular}	{ c | c | c c c | c c c}
	\toprule
     \multirow{2}{2.8cm}{\centering \textbf{Method}} & \multirow{2}{2cm}{\centering \textbf{Image-level Accuracy}} & \multicolumn{3}{c|}{\centering \textbf{Unrelaxed Evaluation}} & \multicolumn{3}{c}{\centering \textbf{Relaxed Evaluation}}\\
     \cline{3-8}
    & & Accuracy & F1 & Jaccard & Accuracy & F1 & Jaccard  \\
     \midrule
    \rowcolor{mygray}
    \multicolumn{8}{c}{\textbf{Cholec80 - Frames $\times$ Frame Rate: 8 $\times$ 4}}\\
    \midrule
    Baseline w/ MA & 89.8 & 90.0 & 84.3 & 75.2 & 90.9 & 88.4 & 79.2\\
    Surgformer w/ MA & 90.4 (\textcolor{red}{+0.6}) &  90.7 (\textcolor{red}{+0.7})& 85.7 (\textcolor{red}{+1.4}) & 76.9 (\textcolor{red}{+1.7}) & 91.6 (\textcolor{red}{+0.7}) & 89.7 (\textcolor{red}{+1.3}) & 80.8 (\textcolor{red}{+1.6}) \\
     \midrule
    \rowcolor{mygray}
    \multicolumn{8}{c}{\textbf{Cholec80 - Frames $\times$ Frame Rate: 12 $\times$ 4}}\\
    \midrule
    Baseline w/ MA & 90.5 & 91.0 & 84.9 & 76.3 & 91.8 & 88.3 & 79.9\\
    Surgformer w/ MA & 90.8 (\textcolor{red}{+0.3}) &  91.1 (\textcolor{red}{+0.1})& 86.3 (\textcolor{red}{+1.4}) & 78.0 (\textcolor{red}{+1.7}) & 92.1 (\textcolor{red}{+0.3}) & 89.9 (\textcolor{red}{+1.6}) & 81.6 (\textcolor{red}{+1.7}) \\
     \midrule
    \rowcolor{mygray}
    \multicolumn{8}{c}{\textbf{Cholec80 - Frames $\times$ Frame Rate: 16 $\times$ 4}}\\
    \midrule
    Baseline w/ MA & 90.5 & 90.6 & 83.3 & 74.3 & 91.7 & 88.7 & 79.3\\
    Surgformer w/ MA & 90.9 (\textcolor{red}{+0.4}) &  91.2 (\textcolor{red}{+0.6})& 85.7 (\textcolor{red}{+2.4}) & 77.6 (\textcolor{red}{+3.3}) & 92.1 (\textcolor{red}{+0.4}) & 89.0 (\textcolor{red}{+0.3}) & 80.8 (\textcolor{red}{+1.5}) \\
    Surgformer w/ KCA & 91.3 (\textcolor{red}{+0.8}) & 91.7 (\textcolor{red}{+1.1})& 86.9 (\textcolor{red}{+3.6}) & 79.1 (\textcolor{red}{+4.8}) & 92.5 (\textcolor{red}{+0.8}) & 89.9 (\textcolor{red}{+1.2}) & 82.1 (\textcolor{red}{+2.8}) \\
    \midrule
    \rowcolor{mygray}
    \multicolumn{8}{c}{\textbf{Autolaparo - Frames $\times$ Frame Rate: 16 $\times$ 4}}\\
    \midrule
    Baseline w/ MA & 83.9 &  84.1 & 70.9 & 62.0 & 85.0 & 74.5 & 66.2  \\
    Surgformer w/ MA & 85.3 (\textcolor{red}{+1.4}) & 85.7 (\textcolor{red}{+1.6}) & 76.9 (\textcolor{red}{+6.0}) & 66.7 (\textcolor{red}{+4.7}) & 86.5 (\textcolor{red}{+1.5}) & 81.9 (\textcolor{red}{+7.4}) & 70.2 (\textcolor{red}{+4.0}) \\
    Surgformer w/ KCA & 85.3 (\textcolor{red}{+1.4}) & 85.5 (\textcolor{red}{+1.4}) & 76.1 (\textcolor{red}{+5.2}) & 65.9 (\textcolor{red}{+3.9}) & 86.4 (\textcolor{red}{+1.4}) & 80.5 (\textcolor{red}{+6.0}) & 70.2 (\textcolor{red}{+4.0}) \\
    \bottomrule
	\end{tabular}}
	\label{tab:ab1}
\end{table*}

\section{Experiments}
\subsection{Experimental Setup}
\label{setup}
\noindent\textbf{Datasets}: To evaluate the performance of our proposed method, we conduct comparison experiments on two public surgical video datasets: Cholec80~\cite{endonet} and Autolaparo~\cite{Autolaparo}. \textbf{\textit{Cholec80}} contains 80 cholecystectomy surgery videos, each annotated with 7 distinct phase labels. The dataset is officially divided into training and test sets, each consisting of 40 videos. \textbf{\textit{Autolaparo}} contains 21 videos of laparoscopic hysterectomy, with manual annotations of 7 surgical phases. Following the official splits, we partition the dataset into 10 videos for training, 4 videos for validation, and the remaining 7 videos for testing. 

\noindent\textbf{Evaluation Metric}: Following the standard settings~\cite{TMRNet,LoViT}, we use four widely-used benchmark metrics: video-level Accuracy, phase-level Precision, phase-level Recall and phase-level Jaccard. In addition, we employ phase-level F1 metric to uniformly measure precision and recall in ablation study $\S~\ref{as}$. We employ both relaxed and unrelaxed evaluation for Cholec80, and unrelaxed evaluation for Autolaparo. In the relaxed evaluation, predictions that fall within a 10-second window around the phase transition and correspond to neighboring phases are considered correct, even if they do not precisely match the ground truth.

\noindent\textbf{Implementation Details}: We initialize the shared weights using corresponding weights pre-trained on Kinetics-400~\cite{TimeSFormer,Kinetics-400_1}, with the remaining layers being randomly initialized. We train the model for 50 epochs with batch size as 24 on 3 NVIDIA GTX 3090 GPUs. We opt AdamW as optimizer, with $\beta_1$ 0.9, $\beta_2$ 0.999, initial learning rate 5e-4, and layer decay 0.75. 
Meanwhile, we set $m=3$ to construct structured temporal segments with lengths of $T/4$, $T/2$ and $T$, respectively. $\alpha$ and $\beta$ are set to $0.5$ for equal blend. Unless stated otherwise, we employ a fixed training configuration with temporal resolution $T$ = 16 and frame rate $R$ = 4 on both Cholec80 and Autolaparo. For testing, we employ $T=16$ on the Autolaparo, and a larger frame length $T=24$ on the Cholec80 for better performance.

\subsection{Ablation Study}
\label{as}
In this section, we conduct in-depth ablation studies to investigate the effectiveness of components on the Cholec80 and Autolaparo datasets. To analyze the contribution of each component, we implement a straightforward baseline by employing TimeSformer~\cite{TimeSFormer} with MA for surgical phase recognition, and systematically introduce each component into the network.

\begin{table*}[t]
    \centering
    \captionof{table}{Overall comparison with the state-of-the-arts on the Cholec80 dataset.}
    \resizebox{10.0cm}{!}{
    \begin{tabular}	{c c c c c c c}
	\toprule
    \multirow{2}{*}{\textbf{Evaluation}} & \multirow{2}{*}{\textbf{Method}} & \multirow{2}{*}{\textbf{Paradigm}} & \textbf{Video-level Metric} & \multicolumn{3}{c}{\textbf{Phase-level Metric}} \\
    \cmidrule(lr){4-4} \cmidrule(lr){5-7}
    & & & \textbf{Accuracy $\uparrow$} & \textbf{Precision $\uparrow$} & \textbf{Recall $\uparrow$} & \textbf{Jaccard $\uparrow$} \\
    \midrule
    \multirow{11}{*}{\textbf{Relaxed}} & EndoNet~\cite{endonet} & Two-stage &
    81.7 $\pm$ 4.2 & 
    73.7 $\pm$ 16.1 &
    79.6 $\pm$ 7.9 &
    - \\
    & MTRCNet-CL~\cite{MTRCNet-CL} & One-stage & 
    89.2 $\pm$ 7.6 & 
    86.9 $\pm$ 4.3 & 
    88.0 $\pm$ 6.9 & 
    - \\
    & PhaseNet~\cite{PhaseNet} &  Two-stage & 
    78.8 $\pm$ 4.7 & 
    71.3 $\pm$ 15.6 & 
    76.6 $\pm$ 16.6 & 
    - \\
    & SV-RCNet~\cite{SV-RCNet} &  One-stage & 
    85.3 $\pm$ 7.3 & 
    80.7 $\pm$ 7.0 & 
    83.5 $\pm$ 7.5 & 
    - \\
    & OHFM~\cite{OHFM} & - & 
    87.3 $\pm$ 5.7 & 
    - & 
    - & 
    67.0 $\pm$ 13.3 \\
    & TeCNO~\cite{TeCNO} & Two-stage & 
    88.6 $\pm$ 7.8 & 
    86.5 $\pm$ 7.0 & 
    87.6 $\pm$ 6.7 & 
    75.1 $\pm$ 6.9 \\
    & TMRNet~\cite{TMRNet} & One-stage & 
    90.1 $\pm$ 7.6 & 
    90.3 $\pm$ 3.3 & 
    89.5 $\pm$ 5.0 & 
    79.1 $\pm$ 5.7 \\
    & Trans-SVNet~\cite{Trans-SVNet} & Two-stage & 
    90.3 $\pm$ 7.1 & 
    90.7 $\pm$ 5.0 & 
    88.8 $\pm$ 7.4 & 
    79.3 $\pm$ 6.6 \\
    & LoViT~\cite{LoViT} & Two-stage & 
    92.4 $\pm$ 6.3 & 
    89.9 $\pm$ 6.1 & 
    90.6 $\pm$ 4.4 & 
    81.2 $\pm$ 9.1 \\
    & SKiT~\cite{SKiT} & Two-stage & 
    \textbf{93.4 $\pm$ 5.2} & 
    90.9 & 
    91.8 & 
    82.6 \\
    & \textbf{Ours} & One-stage & 
    \textbf{93.4 $\pm$ 6.4} & 
    \textbf{91.9 $\pm$ 4.7} & 
    \textbf{92.1 $\pm$ 5.8} & 
    \textbf{84.1 $\pm$ 8.0} \\
    \midrule
    \multirow{5}{*}{\textbf{Unrelaxed}} & Trans-SVNet~\cite{Trans-SVNet} & Two-stage & 
    89.1 $\pm$ 7.0 & 
    84.7 & 
    83.6 & 
    72.5 \\
    & AVT~\cite{AVT} & Two-stage & 
    86.7 $\pm$ 7.6 & 
    77.3 & 
    82.1 & 
    66.4 \\
    & LoViT~\cite{LoViT} & Two-stage & 
    91.5 $\pm$ 6.1 & 
    83.1 & 
    86.5 & 
    74.2  \\
    & SKiT~\cite{SKiT} & Two-stage & 
    \textbf{92.5 $\pm$ 5.1} & 
    84.6 & 
    88.5 & 
    76.7 \\
    & \textbf{Ours} & One-stage & 
    92.4 $\pm$ 6.4
    & \textbf{87.9 $\pm$ 6.9} 
    & \textbf{89.3 $\pm$ 7.8}
    & \textbf{79.9 $\pm$ 10.2} \\
    \bottomrule
	\end{tabular}}
    \label{tab:r1}
\end{table*}

\noindent\textbf{Effectiveness of HTA.} By integrating HTA into the baseline, the variant significantly outperforms the baseline under diverse settings on the Cholec80 dataset. Based on the default training configuration, the variant exhibits significant improvements over the baseline in terms of F1 and Jaccard, surpassing baseline by 2.4\% and 3.3\% in unrelaxed evaluation on the Cholec80. For the baseline, the overall performance degrades as the length increases form 12 to 16, which is attributed to the introduction of irrelevant and noisy information. In contrast to the baseline, the variant gains significant improvements with the increased length, which demonstrates the effectiveness of HTA to learn more discriminative features. The variant demonstrates more significant performance improvements across all metrics when applied to the more challenging Autolaparo. 

\noindent\textbf{Effectiveness of TFA.} By deploying TFA for target-guided aggregation, the variant yields subsequent enhancement across all metrics in both relaxed and unrelaxed evaluation on the Cholec80. We argue that the Cholec80 dataset is relatively straightforward, exhibiting minimal ambiguity during phase transitions. As a result, it benefits from TFA, which prioritizes target frames while effectively suppressing redundant and invalid information. In contrast, the variant suffers a slight performance degradation on the more challenging Autolaparo.

\begin{table*}[t]
\centering
    \captionof{table}{Overall comparison with the state-of-the-arts on the Autolaparo dataset.}
    \resizebox{8.5cm}{!}{
\begin{tabular}	{ c c c c c c}
	\toprule
    \multirow{2}{*}{\textbf{Method}} & \multirow{2}{*}{\textbf{Paradigm}} & \textbf{Video-level Metric} & \multicolumn{3}{c}{\textbf{Phase-level Metric}} \\
    \cmidrule(lr){3-3} \cmidrule(lr){4-6}
    & & \textbf{Accuracy $\uparrow$} & \textbf{Precision $\uparrow$} & \textbf{Recall $\uparrow$} & \textbf{Jaccard $\uparrow$} \\
    \midrule
    SV-RCNet~\cite{SV-RCNet} &  One-stage & 
    75.6 & 
    64.0 & 
    59.7 & 
    47.2 \\
    TMRNet~\cite{TMRNet} & One-stage & 
    78.2 & 
    66.0 & 
    61.5 & 
    49.6 \\
    TeCNO~\cite{TeCNO} & Two-stage & 
    77.3 & 
    66.9 & 
    64.6 & 
    50.7 \\
    Trans-SVNet~\cite{Trans-SVNet} & Two-stage & 
    78.3 & 
    64.2 & 
    62.1 & 
    50.7 \\
    AVT~\cite{AVT} & Two-stage & 
    77.8 & 
    68.0 & 
    62.2 & 
    50.7 \\
    LoViT~\cite{LoViT} & Two-stage & 
    81.4 $\pm$ 7.6 & 
    85.1 & 
    65.9 & 
    56.0 \\
    SKiT~\cite{SKiT} & Two-stage & 
    82.9 $\pm$ 6.8 & 
    81.8 & 
    70.1 & 
    59.9 \\
    \midrule
    \textbf{Ours} & One-stage & 
    \textbf{85.7 $\pm$ 6.9} & 
    \textbf{82.3} & 
    \textbf{75.7} & 
    \textbf{66.7} \\
    \bottomrule
	\end{tabular}}
    \label{tab:r2}
\end{table*}

\begin{figure}[t]
\centering
\includegraphics[width=1.0\linewidth]{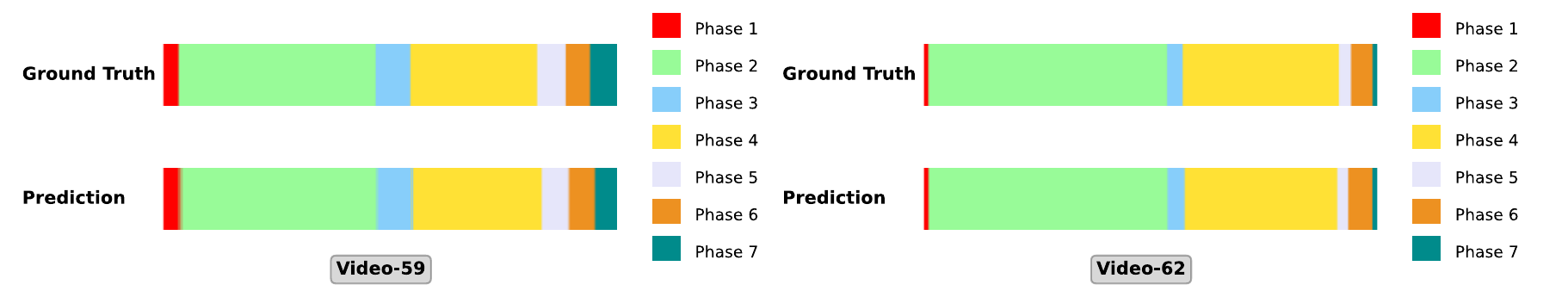}
\caption{Qualitative results on
four video sequences from the
Cholec80 dataset.}
\label{fig:Qualitative Results}
\end{figure}

\subsection{Comparison Results}
We compare the performance of our proposed method with the state-of-the-art methods. Table~\ref{tab:r1} illustrates the results of all compared methods. Our proposed method is comparable with recent state-of-the-art two-stage methods~\cite{LoViT,SKiT} across all metrics in both relaxed and unrelaxed evaluation settings on the Cholec80 dataset. Our proposed Surgformer significantly outperforms the best performance SkiT~\cite{SKiT} on all phase-level metrics, especially in the unrelaxed evaluation. We also present results of more challenging Autolaparo dataset in Table~\ref{tab:r2}. Surgformer also outperforms the best performance SkiT~\cite{SKiT} under all metrics, which achieves gains of 2.8\% and 6.8\% in terms of Accuracy and Jaccard. As shown in Fig.~\ref{fig:Qualitative Results}, we illustrate the phase recognition results of Surgformer on two surgical videos from Cholec80 dataset.
    
\section{Conclusion}
To tackle the challenges associated with spatial-temporal dependency and redundancy in existing methods, we introduce an end-to-end Surgical Transformer (Surgformer), which employs divided spatial-temporal attention and takes a limited set of sparse frames as input. Furthermore, we propose the Hierarchical Temporal Attention (HTA) to capture both global and local information from a target frame-centric perspective. Distinct from conventional temporal attention that primarily emphasizes dense long-range similarity, HTA constructs multiple temporal segments with different temporal resolutions to extract the long-term and short-term information. 
The experimental results on two benchmarks demonstrate that Surgformer learns from sparse frame sequence and outperforms existing two-stage competitors.

\begin{credits}
\subsubsection{\ackname} This work was supported by in part by the Project of Hetao Shenzhen-Hong Kong Science and Technology Innovation Cooperation Zone (HZQB-KCZYB-2020083), the Research Grants Council of the Hong Kong (Project Reference Number: T45-401/22-N), NSFC General Project (62072452), and the Regional Joint Fund of Guangdong under Grant (2021B1515120011).
\end{credits}

%
%
%
\bibliographystyle{splncs04}
\bibliography{Surgformer}

\begin{thebibliography}{10}
\providecommand{\url}[1]{\texttt{#1}}
\providecommand{\urlprefix}{URL }
\providecommand{\doi}[1]{https://doi.org/#1}

\bibitem{ViViT}
Arnab, A., Dehghani, M., Heigold, G., Sun, C., Lu{\v{c}}i{\'c}, M., Schmid, C.: Vivit: A video vision transformer. In: Proceedings of the IEEE/CVF international conference on computer vision. pp. 6836--6846 (2021)

\bibitem{TimeSFormer}
Bertasius, G., Wang, H., Torresani, L.: Is space-time attention all you need for video understanding? In: ICML. vol.~2, p.~4 (2021)

\bibitem{OR1}
Cleary, K., Chung, H.Y., Mun, S.K.: Or2020 workshop overview: operating room of the future. In: International Congress Series. vol.~1268, pp. 847--852. Elsevier (2004)

\bibitem{TeCNO}
Czempiel, T., Paschali, M., Keicher, M., Simson, W., Feussner, H., Kim, S.T., Navab, N.: Tecno: Surgical phase recognition with multi-stage temporal convolutional networks. In: Medical Image Computing and Computer Assisted Intervention--MICCAI 2020: 23rd International Conference, Lima, Peru, October 4--8, 2020, Proceedings, Part III 23. pp. 343--352. Springer (2020)

\bibitem{skill_assessment2}
Dias, R.D., Gupta, A., Yule, S.J.: Using machine learning to assess physician competence: a systematic review. Academic Medicine  \textbf{94}(3),  427--439 (2019)

\bibitem{slowfast}
Feichtenhofer, C., Fan, H., Malik, J., He, K.: Slowfast networks for video recognition. In: Proceedings of the IEEE/CVF international conference on computer vision. pp. 6202--6211 (2019)

\bibitem{OR2}
Franke, S., Rockstroh, M., Hofer, M., Neumuth, T.: The intelligent or: design and validation of a context-aware surgical working environment. International Journal of Computer Assisted Radiology and Surgery  \textbf{13},  1301--1308 (2018)

\bibitem{Trans-SVNet}
Gao, X., Jin, Y., Long, Y., Dou, Q., Heng, P.A.: Trans-svnet: Accurate phase recognition from surgical videos via hybrid embedding aggregation transformer. In: Medical Image Computing and Computer Assisted Intervention--MICCAI 2021: 24th International Conference, Strasbourg, France, September 27--October 1, 2021, Proceedings, Part IV 24. pp. 593--603. Springer (2021)

\bibitem{AVT}
Girdhar, R., Grauman, K.: Anticipative video transformer. In: Proceedings of the IEEE/CVF international conference on computer vision. pp. 13505--13515 (2021)

\bibitem{SV-RCNet}
Jin, Y., Dou, Q., Chen, H., Yu, L., Qin, J., Fu, C.W., Heng, P.A.: Sv-rcnet: workflow recognition from surgical videos using recurrent convolutional network. IEEE transactions on medical imaging  \textbf{37}(5),  1114--1126 (2017)

\bibitem{MTRCNet-CL}
Jin, Y., Li, H., Dou, Q., Chen, H., Qin, J., Fu, C.W., Heng, P.A.: Multi-task recurrent convolutional network with correlation loss for surgical video analysis. Medical image analysis  \textbf{59},  101572 (2020)

\bibitem{TMRNet}
Jin, Y., Long, Y., Chen, C., Zhao, Z., Dou, Q., Heng, P.A.: Temporal memory relation network for workflow recognition from surgical video. IEEE Transactions on Medical Imaging  \textbf{40}(7),  1911--1923 (2021)

\bibitem{Kinetics-400_1}
Kay, W., Carreira, J., Simonyan, K., Zhang, B., Hillier, C., Vijayanarasimhan, S., Viola, F., Green, T., Back, T., Natsev, P., et~al.: The kinetics human action video dataset. arXiv preprint arXiv:1705.06950  (2017)

\bibitem{skill_assessment1}
Kowalewski, K.F., Garrow, C.R., Schmidt, M.W., Benner, L., M{\"u}ller-Stich, B.P., Nickel, F.: Sensor-based machine learning for workflow detection and as key to detect expert level in laparoscopic suturing and knot-tying. Surgical endoscopy  \textbf{33},  3732--3740 (2019)

\bibitem{LoViT}
Liu, Y., Boels, M., Garcia-Peraza-Herrera, L.C., Vercauteren, T., Dasgupta, P., Granados, A., Ourselin, S.: Lovit: Long video transformer for surgical phase recognition. arXiv preprint arXiv:2305.08989  (2023)

\bibitem{SKiT}
Liu, Y., Huo, J., Peng, J., Sparks, R., Dasgupta, P., Granados, A., Ourselin, S.: Skit: a fast key information video transformer for online surgical phase recognition. In: Proceedings of the IEEE/CVF International Conference on Computer Vision. pp. 21074--21084 (2023)

\bibitem{PhiTrans}
Liu, Y., Zhong, X., Zhai, S., Du, Z., Gao, Z., Huang, Q., Zhang, C.Y., Jiang, B., Pandey, V.K., Han, S., et~al.: Prompt-enhanced hierarchical transformer elevating cardiopulmonary resuscitation instruction via temporal action segmentation. Computers in Biology and Medicine  \textbf{167},  107672 (2023)

\bibitem{optimization}
Neumuth, T.: Surgical process modeling. Innovative surgical sciences  \textbf{2}(3),  123--137 (2017)

\bibitem{PhaseNet}
Twinanda, A.P., Mutter, D., Marescaux, J., de~Mathelin, M., Padoy, N.: Single-and multi-task architectures for surgical workflow challenge at m2cai 2016. arXiv preprint arXiv:1610.08844  (2016)

\bibitem{endonet}
Twinanda, A.P., Shehata, S., Mutter, D., Marescaux, J., De~Mathelin, M., Padoy, N.: Endonet: a deep architecture for recognition tasks on laparoscopic videos. IEEE transactions on medical imaging  \textbf{36}(1),  86--97 (2016)

\bibitem{Autolaparo}
Wang, Z., Lu, B., Long, Y., Zhong, F., Cheung, T.H., Dou, Q., Liu, Y.: Autolaparo: A new dataset of integrated multi-tasks for image-guided surgical automation in laparoscopic hysterectomy. In: International Conference on Medical Image Computing and Computer-Assisted Intervention. pp. 486--496. Springer (2022)

\bibitem{AMCNet}
Yang, S., Zhang, L., Qi, J., Lu, H., Wang, S., Zhang, X.: Learning motion-appearance co-attention for zero-shot video object segmentation. In: Proceedings of the IEEE/CVF international conference on computer vision. pp. 1564--1573 (2021)

\bibitem{OHFM}
Yi, F., Jiang, T.: Hard frame detection and online mapping for surgical phase recognition. In: Medical Image Computing and Computer Assisted Intervention--MICCAI 2019: 22nd International Conference, Shenzhen, China, October 13--17, 2019, Proceedings, Part V 22. pp. 449--457. Springer (2019)

\bibitem{FVOS}
Zhang, L., Lin, Z., Zhang, J., Lu, H., He, Y.: Fast video object segmentation via dynamic targeting network. In: Proceedings of the IEEE International Conference on Computer Vision. pp. 5582--5591 (2019)

\bibitem{UVOSHT}
Zhang, L., Zhang, J., Lin, Z., M{\v{e}}ch, R., Lu, H., He, Y.: Unsupervised video object segmentation with joint hotspot tracking. In: Proceedings of the European Conference on Computer Vision. pp. 490--506. Springer International Publishing (2020)

\bibitem{MATNet}
Zhou, T., Li, J., Wang, S., Tao, R., Shen, J.: Matnet: Motion-attentive transition network for zero-shot video object segmentation. IEEE Transactions on Image Processing  \textbf{29},  8326--8338 (2020)

\end{thebibliography}
\end{document}